\pdfoutput=1
\documentclass[11pt]{article}

\usepackage[]{ACL2023}

\usepackage{times}
\usepackage{latexsym}

\usepackage[T1]{fontenc}
\usepackage[utf8]{inputenc}

\usepackage{microtype}

\usepackage{inconsolata}

\usepackage{CJKutf8}
\newcommand{\chinese}[1]{\begin{CJK}{UTF8}{gkai}{}#1\end{CJK}}

\usepackage{algorithm}
\usepackage{algorithmicx}
\usepackage{algpseudocode}
\usepackage{setspace}
\usepackage{graphicx} %use graph format
\usepackage{epstopdf}
\usepackage{bbding}
\usepackage{color}
\usepackage{makecell}
\usepackage{booktabs}
\usepackage{multirow}
\usepackage{float}
\usepackage{enumerate}
\usepackage{fancyhdr}
\usepackage{bm}

\newcommand{\PreserveBackslash}[1]{\let\temp=\\#1\let\\=\temp}
\newcolumntype{C}[1]{>{\PreserveBackslash\centering}p{#1}}
\newcolumntype{R}[1]{>{\PreserveBackslash\raggedleft}p{#1}}
\newcolumntype{L}[1]{>{\PreserveBackslash\raggedright}p{#1}}

\usepackage{amsmath}
\usepackage{amsfonts}
\usepackage{hyperref}

\title{Lexical Complexity Controlled Sentence Generation}

\author{
    Jinran Nie\textsuperscript{1},
    Liner Yang\textsuperscript{1}, 
    Yun Chen\textsuperscript{2},
    Cunliang Kong\textsuperscript{1},
    Junhui Zhu\textsuperscript{1},
    Erhong Yang\textsuperscript{1} \\
    \textsuperscript{1} School of Information Science, Beijing Language and Culture University \\
    \textsuperscript{2} School of Information Management and Engineering, \\ Shanghai University of Finance and Economics
}

\begin{document}
\maketitle
\begin{abstract}
Text generation rarely considers the control of lexical complexity, which limits its more comprehensive practical application. We introduce a novel task of lexical complexity controlled sentence generation, which aims at \textit{keywords to sentence} generation with \textit{desired complexity levels}. It has enormous potential in domains such as grade reading, language teaching and acquisition. The challenge of this task is to generate fluent sentences only using the words of given complexity levels. We propose a simple but effective approach for this task based on complexity embedding. Compared with potential solutions, our approach fuses the representations of the word complexity levels into the model to get better control of lexical complexity. And we demonstrate the feasibility of the approach for both training models from scratch and fine-tuning the pre-trained models. To facilitate the research, we develop two datasets in English and Chinese respectively, on which extensive experiments are conducted. Results show that our approach better controls lexical complexity and generates higher quality sentences than baseline methods.
\end{abstract}

\section{Introduction}
%\noindent 

% Controlling lexical complexity in sentence generation has a wide range of applications in language learning and grade reading. In the fields of language teaching and acquisition, language instructors need to make teaching materials with example sentences, either synthetically designed or from authentic resources \cite{caro2017lexis, lu2019sentence}. Language instructors and textbook compilers are usually required to create example sentences that only use the words at specific complexity for students in specific levels \cite{nordlund2020vocabulary, laufer2021lexical}, which is very time-consuming and exhausting. By controlling the complexity of words, the Controllable Text Generation (CTG) task can provide support for educators and language learners to explore, analyze, and select proper example sentences. Besides, it can also assist in developing grade reading materials \cite{ryu2020analysis, al2021efl, amer2021lexical}.
Controlling lexical complexity in sentence generation has a wide range of applications in language learning and grade reading. In the fields of language teaching and acquisition, language instructors and textbook compilers need to make teaching materials with example sentences, either synthetically designed or from authentic resources \cite{caro2017lexis, lu2019sentence}. In most cases, they are required to create example sentences that only use the words at particular complexity for students in specific levels \cite{nordlund2020vocabulary, laufer2021lexical}, which is very time-consuming and exhausting. By controlling the complexity of words, the Controllable Text Generation (CTG) task can support educators and language learners to explore, analyze, and select proper example sentences. Besides, it can also assist in the development of graded reading materials \cite{ryu2020analysis, al2021efl, amer2021lexical}.

Controllable text generation, a significant area of natural language generation, contains a series of tasks that aim to generate text according to the given controlled requirements \cite{prabhumoye2020exploring,zhang2022survey}. CTG systems usually focus on controlling text attributions such as sentiment \cite{hu2017toward, zhang2019emotional, samanta2020fine}, topic \cite{dathathri2019plug, tang2019topic, khalifa2020distributional} or keywords \cite{he2021parallel, zhang2020pointer, mcmc-xlnet}, generating poems or couplets with specific formats \cite{chen2019sentiment, shao2021sentiment, sheng2021songmass}, and even predicting descriptions from structured data \cite{zhao2020bridging, su2021plan, ribeiro2021structural}. However, few works have been devoted to strict control over the lexical complexity for text generation. Although lexical simplification has been paid attention to the text simplification task through substitution \cite{kriz2018simplification}, it cannot strictly control the lexical complexity levels of the generated sentence.

\begin{figure}[t]
\centering
\includegraphics[width=6cm]{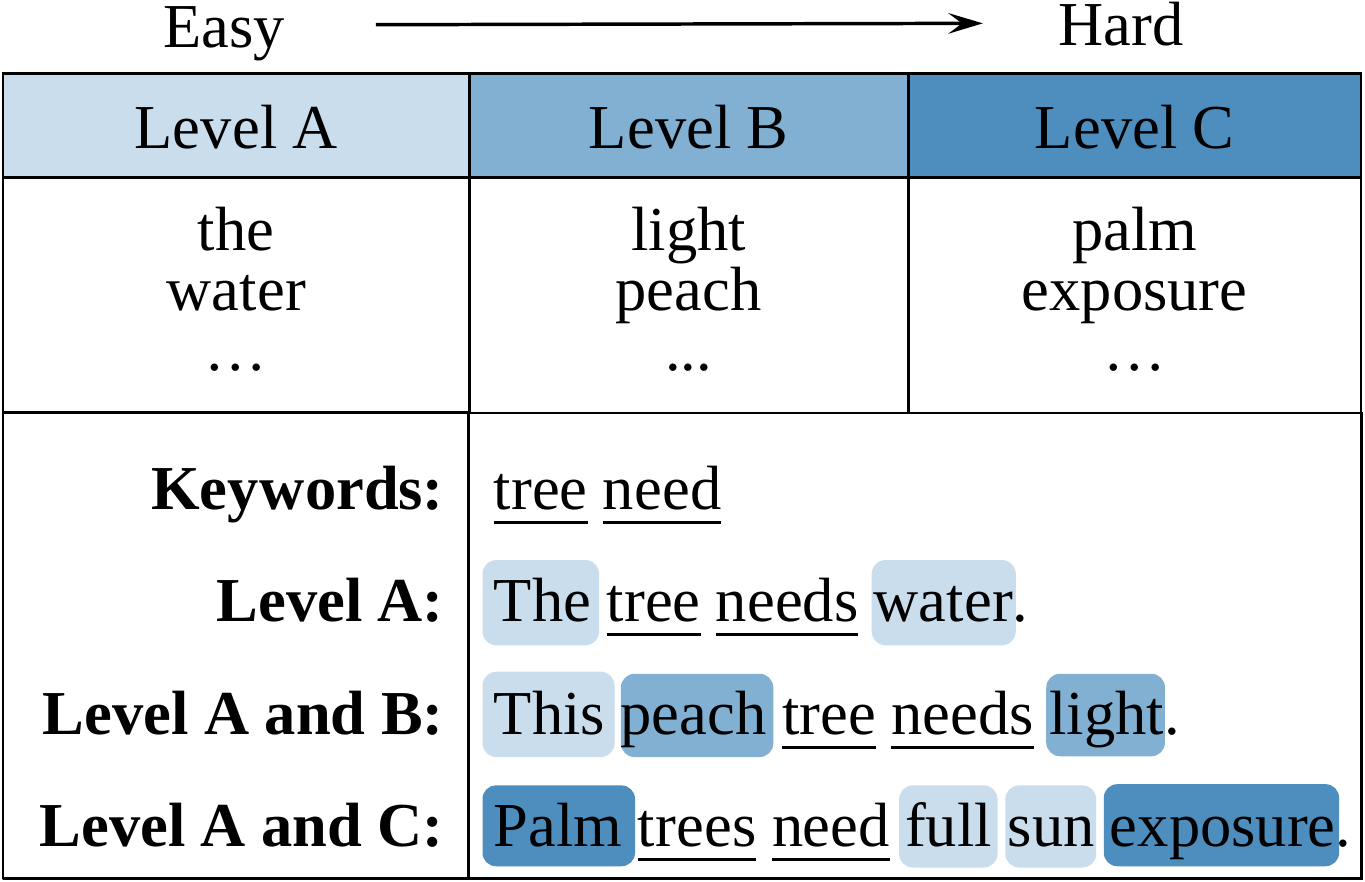}
\caption{An example for lexical complexity controlled sentence generation. There are three complexity levels (A, B, and C) from easy to hard. Given the keywords ``tree'' and ``need'', we will generate ``The tree needs water.'' if required to use all words from level A and generate ``This peach tree needs light.'' if required to use words from both level A and B as both “peach” and “light” are in level B.}
\label{figure1}
\end{figure}

\begin{figure*}[t]
\centering
\includegraphics[width=14cm]{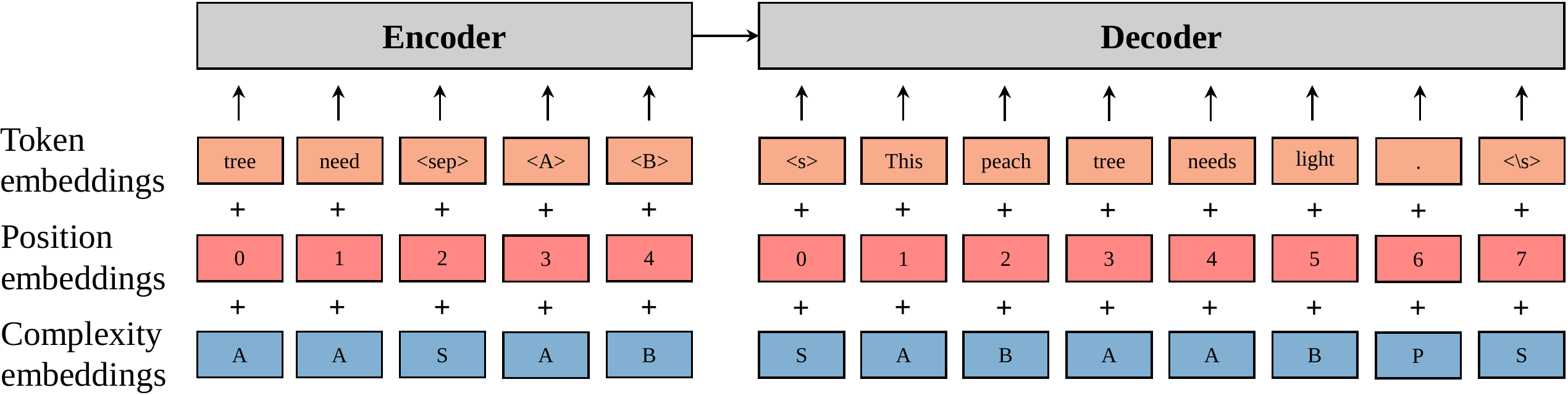}
\caption{
    Encoder-Decoder model with our proposed CE method.
    The representation of each input token is a summary of three embeddings, which are token embedding, position embedding, and complexity embedding.
    And we concatenate the keywords and complexity level tokens as the input sequence of the encoder.
    Note that the special tokens correspond to the complexity level of ``S'', and the punctuation correspond to ``P''.
}
% \caption{Encoder-Decoder model with complexity embedding. In the model, the token embeddings and position embeddings are the same as the standard Transformer. In complexity embeddings, the words of different complexity levels use different complexity tokens. For example, the complexity level of the word ``tree'' is A, and ``peach'' is B. The word tokens ``$\langle A \rangle$'' and ``$\langle B \rangle$'' are in accordance with the complexity tokens. The special tokens like ``$\langle sep \rangle$'' and ``$\langle s \rangle$'' use S to represent their complexity and the punctuation uses P to represent the complexity.}
\label{figure4}
\end{figure*}

To this end, we propose a new task of lexical complexity controlled sentence generation, which requires that keywords and  complexity levels be given to generate a sentence including the keywords and consisting of the words in the given complexity levels. For example, as shown in Figure \ref{figure1}, we assume that there are three complexity levels (A, B, and C) from easy to hard. Given the keywords, we can generate sentences consisted with words of different complexity according to the given levels.

It is challenging to generate fluent sentences for given keywords while using the words only at specific complexity levels. This can be regarded as an extension and a particular case of lexical CTG task \cite{mcmc-xlnet, miao2019cgmh, zhang2020pointer}. Differently, it combines two aspects of constraints during generation: keywords constraint the semantics, and lexical complexity levels constraint the surface form. It is difficult for the model to select suitable words from a specific subspace satisfying the above two constraints in each generation process. We formulate this problem in Section \ref{define}. 

Some previous works can be customized as solutions to this problem, which are divided into three branches: controlled decoding, prompting, and reranking. The first method forces to change the probability distribution during the decoding phase to ensure that only words of the specified levels are used in the generation \cite{dathathri2019plug, DBA}. But the hard constraint may lead to poor quality generation quality. The second one considers lexical complexity through prompting \cite{brown2020language, raffel2020exploring, liang2021prefix} in the input of the model, which introduce coarse grained information of training and inference. The method of reranking is to select the sentence that best meets the lexical complexity requirements from the candidates \cite{ravaut2022summareranker, pandramish2020checkpoint}, which executes after decoding and does not consider lexical complexity in the training time.

The complexity constraint requires models to aware of lexical complexity and respond to complexity control signals. Therefore, we use two mechanisms  as enhancements to the transformer-based models. \textit{For the complexity awareness}, we propose the Complexity Embedding (\textbf{CE}) method, which represents the complexity levels with trainable embeddings. We incorporate the CEs into both training and prediction processes by fusing the CEs and word embeddings as token representations, which is simple but effective. \textit{For responding to complexity control signals}, we concatenate special tokens corresponding to specific complexity levels with the keywords as the input sequence. To combine the awareness and response, we use CEs to represent these special tokens. The experiments show that our proposed method is effective for both training from scratch and fine-tuning the pre-trained language models. And compared to the baseline methods, our method achieves significant improvement in the restriction of lexical complexity levels and generation quality. Our main contributions include:
\begin{itemize}
\item We propose a new task of lexical complexity controlled sentence generation and two datasets in English and Chinese for this task. To evaluate the satisfaction of the lexical complexity constraint, we develop four metrics.

\item We propose a new method for this task based on complexity embedding.

% \item We propose two types of evaluation metrics for lexical complexity level constraints including four metrics.
% accuracy rate at the lexical level; precision rate, recall rate and F1 at the lexical complexity level.

\item A series of baseline methods are implemented for this task and experimental results show that the complexity embedding method we propose significantly outperforms the baseline methods.

%, including three types of methods: controlled decoding methods, reordering methods, and prompt methods.
\end{itemize}

\section{Problem Definition} \label{define}

% \subsection{Problem Definition} \label{define}
\textbf{Lexical Complexity Sentence Generation} aims at keywords to sentence generation with desired complexity levels. First, we give the keywords set $K=\{k_1,k_2,...,k_m\}$ and the complexity levels $L=\{l_1,l_2,...,l_n\}$ which correspond to a subset $D=\{W_1\cup W_2 \cup ...\cup W_n\}$ of the whole vocabulary $V$ and $W_i$ is the word set of complexity level $l_i$. The control elements in this task include three parts:

First, we define a predicate $F(K,Y)$ to be a boolean function indicating the occurrence of keyword $k_i$ in a generated sequence $Y=y_1, y_2,...,y_t$, and $t$ is the sequence length.
\begin{align}
    C_1 &= F(K, Y) \\
    F(K, Y) &\equiv \forall \ i, k_i \in Y
\end{align}
where $C_1$ is the keywords constraint which means the keywords are required to be included in the generated sentence.

Second, we define a predicate $G(Y,D)$ to be a boolean function indicating the occurrence of a word  $y_i$ which is a word of the sentence $Y$ in a word set $D$. 
\begin{align}
    C_2 &= G(Y, D) \\ \label{equation:c2}
    G(Y, D) &\equiv \forall \ i, y_i \in D
\end{align}
where $C_2$ is the complexity constraint on word which means the words in the generated sentence are required to be the words of the given complexity levels.

Then, we define a predicate $H(Y, W_i)$ to be a boolean function indicating that there exist at least one word in the generated sentence in the $W_i$.
\begin{align}
    C_3 = H(Y, W_1)& \wedge H(Y, W_2) ... \wedge H(Y, W_n)\\
    H(Y, W_i)& \equiv \exists \ j, y_j \in W_i
\end{align}
where $C_3$ is the constraint on the species of complexity level which means the lexical levels of the generated sentence need cover all the given levels. 
% \begin{equation}
% \label{formula1}
% \begin{aligned}
% C_1&=\left\{ F(k_1, Y)\wedge F(k_2, Y)...\wedge F(k_m, Y) \right\} \\ 
% C_2&= \left\{ G(y_1, D)\wedge G(y_2, D)...\wedge G(y_t, D) \right\} \\ 
% C_3&=\left\{ G(y_{i_1}, W_1)\wedge G(y_{i_2}, W_2)...\wedge G(y_{i_n}, W_n) \right\}
% \end{aligned}
% \end{equation}
% where $C_1$ is the keywords constraint which means the keywords are required to be included in the generated sentence, $C_2$ is the complexity constraint on word which means the words in the generated sentence are required to be the words of the given complexity levels, and $C_3$ is the constraint on the species of complexity level which means the lexical levels of the generated sentence need cover all the given levels. 

The task requires to seek optimal sequences in which all constraints are satisfied as much as possible. The formula is as follows:
\begin{equation}
\hat{Y} = \mathop{\rm{arg\,max}}\limits_{Y \in \mathcal{Y}}\log P_\theta\big(Y|K,L \big) \ \ \ \text{where} \ \ \sum\limits_{i=1}^{N}C_i=N %\nonumber
\end{equation}
% \begin{equation}
% \begin{aligned}
% &\left\{ F(k_1, Y)\vee F(k_2, Y)...\vee F(k_m, Y) \right\} \vee \Rightarrow \\ 
% &\left\{ G(y_{i_1}, L_1)\vee G(y_{i_2}, L_2)...\vee G(y_{i_n}, L_n) \right\} \vee \\ 
% &\left\{ G(y_1, D)\vee G(y_2, D)...\vee G(y_t, D) \right\}
% \end{aligned}
% \end{equation}
% \begin{equation}
% Y = \mathop{\rm{arg\,max}}\limits_{y_t \in D}\log p_\theta\big(y_t|K,L,y_{<t} \big)\nonumber
% \end{equation}
% where $Y$ is the sentence is made up of words in the given complexity levels and it is containing the given keywords. Sequence-to-sequence generation is the task of generating an output sequence given an input sequence. We consider standard left-to-right of autoregressive models, thus the predicted tokens $y_{\leq t}$ before current step $t$ will be given as the conditions to predict $y_t$.

% In this section, We propose a novel method for lexical complexity controlled sentence generation, which is based on complexity embedding. And we also present several approaches for lexical complexity controlled sentence generation, including the method of controlling in the decoding stage, prompting method and re-ranking approach.

\section{Method}
% We propose lexical complexity embedding method (CE) for lexical complexity controlled sentence generation.
As illustrated in Figure \ref{figure4}, our model is based on the encoder-decoder architecture.
To make the model aware of the complexity levels, we fuse the complexity into the task by designing a lexical complexity embedding for each token.
To make the model respond to specific complexity levels, we insert special tokens corresponding to complexity levels into the input sequence as controllable elements.
This section introduces these two key components as well as the training and inference strategy.

\subsection{Complexity Embedding}
% We use encoder-decoder architecture which can be the standard Transformer model or Pre-trained model. 
%  There is two pattern for complexity embedding: from scratch training on Transformer and fine-tune on pre-trained model. Because there is no complexity embedding layer in pre-trained model, we will add a new complexity embedding layer when fine-tune on the pre-trained model.
% As illustrated in Figure \ref{figure4}, our model is an encoder-decoder architecture, and the input is the concatenation of the keywords and the complexity levels.
We initialize a learnable matrix $\mathbf{M} \in \mathbb{R}^{U\times dim}$ as representations of complexity levels, where $U$ is the total number of complexity levels, and $dim$ is the dimensions of each embedding.
For each token input to the encoder and decoder, we retrieve a predefined hash-table to obtain its complexity level $l_i$.
Then we get the corresponding complexity embedding by $com_i = \mathbf{M}_i$.
The final embedding of this token $emb_i$ is as following:
\begin{equation}
    emb_i = tok_i + pos_i + com_i
\end{equation}
where $tok_i$ and $pos_i$ are token and positional embeddings, which are obtained according to \citet{Transformer}.

% The input of the model also includes token embeddings, position embeddings and complexity embeddings. For the sake of illustration, we assume that the complexity levels are divided into three grades (A, B and C) in Figure 2. 
For example, as shown in Figure \ref{figure4}, when two keywords ``tree'' and ``need'' along with two complexity levels A and B are required, the sentence ``This peach tree needs light.'' is generated which satisfies both constraints. We use different complexity representations (mapping into a complexity embedding) for words of different complexity levels. And the complexity representations of special tokens and punctuation are also different.

In practice, we apply the BPE (byte pair encoding) \cite{sennrich2015neural} algorithm to split words into sub-word tokens to mitigate the OOV (out-of-vocabulary) problem. We mark each sub-word with the same complexity level as the original word.
More details about the complexity levels can be found in the Appendix \ref{EmbID}.

% \begin{equation}
% Emb_i=\left\{
% \begin{aligned}
% &KEY_{layer}(x_i), \ \ i \le \lvert K \rvert \\
% &Layer_2(x_i), \ \ otherwise
% \end{aligned}
% \right.
% \end{equation}

% The complexity id of words in A level is 1 and B level is 2. Thus the complexity embeddings of words ``This, tree, needs, water'' are 1 and the complexity embedding of word ``peach'' is 2. And for punctuation we use 0 as its complexity embedding such as the ``.'' in the example. Note that, in the input, we use the same words complexity id to present the special token of complexity level. For example, the complexity embedding id of level A (``$\left \langle A \right \rangle$'') is the same with the complexity id of words in level A. In the input sequence, the keywords and complexity levels are concatenate by the special token ``$\left \langle sep \right \rangle$''.
 
% Another issue for model training with complexity embedding is that the difficulties of subword tokens in byte pair encoding are not ensured to appear in the vocabulary complexity level. Here we give the subword tokens the complexity level before subword segmentation. 
 
\subsection{Controllable Elements}
As illustrated in Equation \ref{equation:c2}, each word in the sentence $Y$ is constrained to the word set $D$.
To achieve this, we design a set of special tokens $Z=\{z_1, z_2, \dots, z_n\}$, where each token corresponds to a complexity level in $L$.
% And then we form the set $Z^{'}$ consists of special tokens corresponding to the complexity level of each word in the sentence $Y$.
% \begin{align}
%     Z^{'} &= \{\ComTok(y_i)\}_{i=1}^{t} \\
%     z_j &= \ComTok(y_i)
% \end{align}
% where $\ComTok(\cdot)$ is a function to obtain the complexity token corresponding to a word.

We concatenate the keywords and the special tokens as the input sequence $X=[K; \langle sep \rangle; Z]$.
And we refer the special tokens $Z$ as controllable elements, as they control the complexity of the generated sentence.
Note that the complexity embedding of $z_i$ is that of the level $l_i$.

\subsection{Training and Inference}
We train the complexity embedding in the Transformer model from scratch or fine-tune the pre-trained model discriminatively as there is no complexity embedding layer in the pre-trained process. If a model is trained from scratch, the parameters of complexity embedding will be trained the same as other parameters in the model. If the complexity embedding is added to a pre-trained model for fine-tuning, we first train the complexity embedding layer by fixing the original parameters of the pre-trained model and then fine-tune the whole model.

During the training process, in fact, both the word embedding and the complexity embedding are in a teach-forcing pattern through the ground truth. At the time of inference, the next word embedding at each step will be predicted by the probability distribution of the vocabulary of the model. Since the complexity level of the next word is unknown at each step of the inference stage, we utilize a look-up table method to map the predicted token id to complexity id. The table is a mapping relation between the token id and its complexity id on the whole vocabulary. At each step, the token id will be predicted by the model. We get its complexity id through its token id and the table. The complexity id and token id will then be given as the input for the next step of inference.

% \begin{figure}[t]
% \centering
% \includegraphics[width=7cm]{./images/figure3.pdf}
% \caption{The coverage rate of words in two dataset on the vocabulary of complexity level.}
% \label{figure2}
% \end{figure}

\begin{figure}[t]
\centering
\includegraphics[width=8cm]{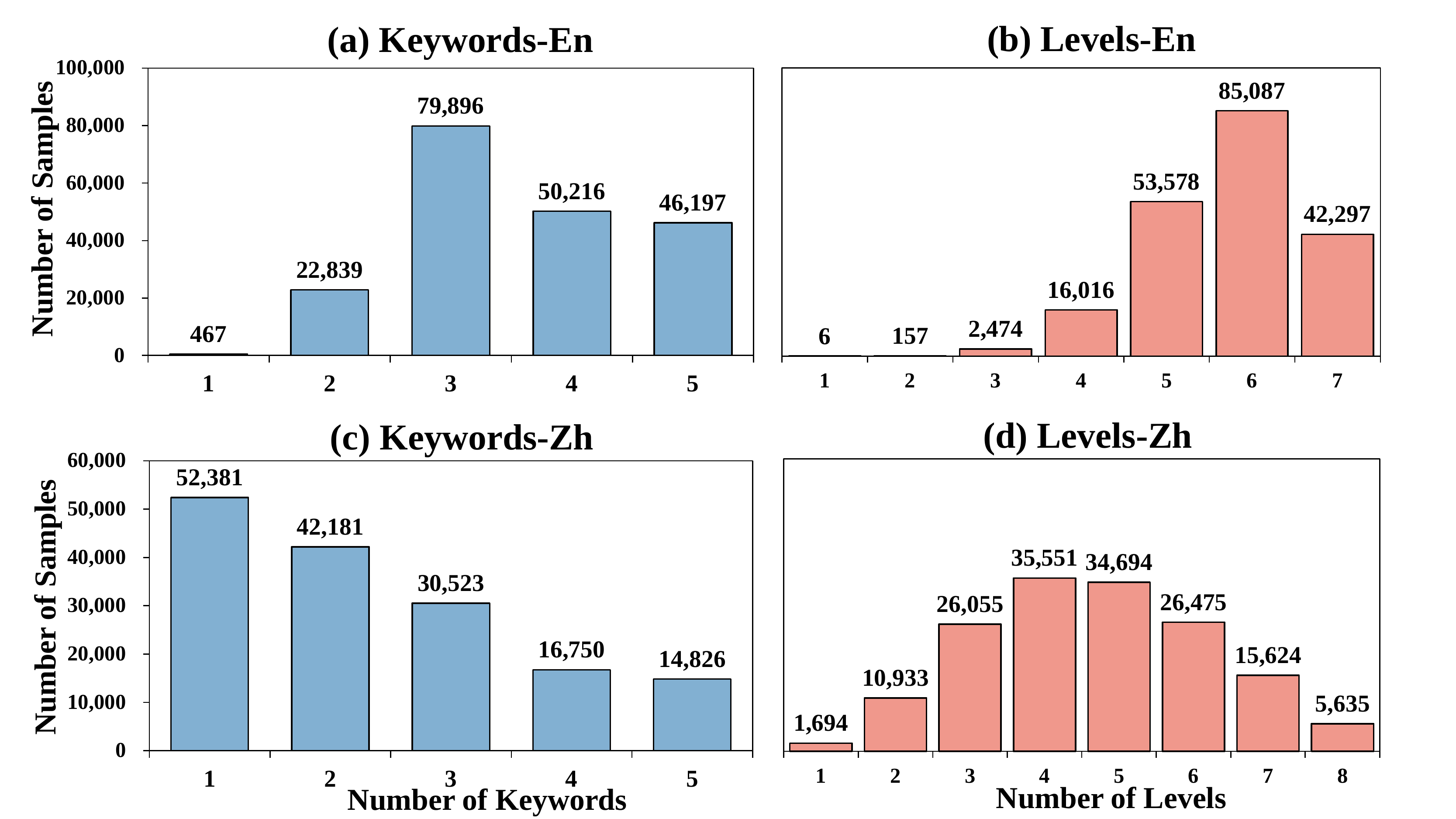}
\caption{Distributions of the number of keywords and complexity levels.}
\label{figure3}
\end{figure}

\begin{table}[t]
	\small
	\centering
	\renewcommand\arraystretch{1.5}
	\begin{tabular}{C{1cm}|C{1cm}|C{1cm}|C{1cm}|C{1cm}}
		\Xhline{1pt}
		%\hline
		Dataset & Train & Valid & Test & Total \\
		%\Xhline{1pt}
		\Xhline{1pt}
		English & 180,000 & 16,000 & 3,615 &  199,615\\
		\hline
		Chinese & 140,000 & 14,000 & 2,661 & 156,661 \\
		\Xhline{1pt}
	\end{tabular}
	\caption{Statistics of the two datasets. }
	\label{tabel1}
\end{table}

\section{Datasets and Evaluation Metrics} \label{data-metric}

% We now describe our datasets and evaluation metrics. We present two datasets for lexical complexity controlled sentence generation in English and Chinese and then analyze both from word coverage and the number of keywords and complexity levels distributions. The evaluation metrics include two aspect: generation quality and satisfaction of constraint.
% We evaluate the performance from general text generation metrics and specific metrics for controlling the keywords and the complexity of words.
% We choose the news corpus in English and the second language textbook corpus in Chinese to cover different fields. To guarantee the word complexity requirement, we need to filter and process the raw data.

\subsection{Dataset Construction}
We present two datasets for lexical complexity controlled sentence generation in English and Chinese. The English raw corpus is collected from the monolingual English News dataset in ACL2019 WMT. The Chinese raw corpus is collected from 500 textbooks for Chinese L2 learners. We adopt the English word complexity levels in the Common European Framework of Reference for Languages (CEFR) \footnote{https://www.englishprofile.org/wordlists/evp} which is divided into six complexity levels (A1, A2, B1, B2, C1, and C2). The word complexity levels in Chinese Proficiency Grading Standards for International Chinese Language Education (CPGS) \footnote{http://www.chinesetest.cn} is divided into seven complexity levels (1 to 7). The process for cleaning data is divided into three steps: split the raw data into sentences and choose the proper sentences; obtain the keywords from the sentences; get the lexical complexity levels from the sentences. The statistics of the two datasets are in the Tabel \ref{tabel1}. More details are in the Appendix \ref{dataDetails}.

% For extensive and practical application, the English dataset is collected from news, and the Chinese dataset is from the material for second language learners.

% We adopt the English word complexity levels in Common European Framework of Reference for Languages (CEFR) \footnote{https://www.englishprofile.org/wordlists/evp} which is actually divided into six complexity levels (A1, A2, B1, B2, C1, and C2). In Chinese, the word complexity levels in Chinese Proficiency Grading Standards for International Chinese Language Education (CPGS) \footnote{http://www.chinesetest.cn} which is divided into six complexity levels (1 to 7). For extensive and practical application, our English dataset is collected from news and Chinese dataset is from the material for second language learners.
\subsection{Analysis of the Datasets}

\subsubsection{Coverage of Words with Levels}
We first analyze the two datasets from the coverage rate of complexity level vocabulary. Due to the requirement of complexity level, the target text is proper to cover most of the vocabulary of complexity level. Both of the two datasets have covered over 93\% of the vocabulary of complexity levels. 

\subsubsection{Distributions of the Number of Keywords and Complexity Levels}
One or multiple complexity levels and keywords are given as the input to generate sentences. We give the distribution of the number of keywords and the complexity levels in Figure \ref{figure3}. From the statistics of (a) and (c) in Figure \ref{figure3}, the number of keywords in all samples has covered the range of 1 to 5 both in the English and Chinese datasets, but the distributions are quite different. On account of the average sentence length of English news data is longer than the Chinese corpus, the number of keywords in English is larger.  From the statistics in (b) and (d) of Figure \ref{figure3}, the number of complexity levels distribution of the Chinese dataset is close to a standard normal distribution, and the English dataset concentrates on a wider range of complexity levels. This indicates that in the English dataset it tends to use more words of different complexity levels in the same sentence.
% \subsubsection{coverage rate of complexity level vocabulary} The vocabulary of complexity level 
% \subsubsection{distribution of complexity level}
% \subsubsection{keywords number distribution}

\begin{table*}[t]
	\small
	\centering
	\renewcommand\arraystretch{1.2}
	\begin{tabular}{L{2cm}|R{0.7cm}R{0.7cm}|R{0.7cm}R{0.7cm}|C{1.6cm}|R{0.7cm}R{0.7cm}|R{0.7cm}R{0.7cm}|R{1cm}}
		\Xhline{1pt}
		\multirow{2.5}{*}{Metrics} & \multicolumn{2}{c|}{\ \ \ \ BLEU(\%)} & \multicolumn{2}{c|}{\ \ \ \ NIST(\%)} & \multirow{2.5}{*}{METEOR(\%)} & \multicolumn{2}{c|}{\ \ \ Entropy(\%)} & \multicolumn{2}{c|}{\ \ \ Distinct(\%)}  &  \multirow{2.5}{*}{PPL}  \\[3pt]
		& B-2 & B-4  & N-2 & N-4  &  & E-2 & E-4  & D-1 & D-2 &    \\[3pt]
		\Xhline{1pt}
		 \multicolumn{11}{l}{\textbf{Training Transformer from scratch}} \\
% 		\Xhline{1pt}
        \hline
        K2S & 16.58 & 4.57 & 3.14 & 3.27 & 15.23 & 8.20 & 10.23 &  \textbf{5.93} & 24.76 & 74.91 \\
		\hline
		Ctrl-decoding & 12.12 & 3.16 & 2.45 & 2.61 & 11.72 & 7.28 & 9.22 &  5.27 & 20.14 & 286.50 \\
		Prompting & 18.19 & 5.73 & 3.57 & 3.64 & 15.93 & 8.30 & 10.36 &  6.10 & 25.55 & 52.10 \\
		Reranking & \textbf{18.47} & 6.27 & 3.52 & 3.60 & 15.99 & 7.87 & 9.79 &  5.93 & 22.70 & 47.81 \\
% 		\Xhline{1pt}
        \hline
		CE (ours) & 18.37 & \textbf{6.66 }& \textbf{3.64} & \textbf{3.69} & \textbf{16.06} & \textbf{8.43} & \textbf{10.47} &  5.80 & \textbf{25.75} & \textbf{42.06} \\
		\Xhline{1pt}
        \multicolumn{11}{l}{\textbf{Fine-tuning BART}} \\
        \hline
		K2S & 17.40 & 5.96 & 3.20 & 3.26 & 15.60 & 8.60 & 10.52 &  6.36 & 28.53 & 33.11 \\
		\hline
		Ctrl-decoding & 14.17 & 3.55 & 2.73 & 2.48 & 13.15 & 8.03 & 9.87&  5.96 & 21.96 & 223.43 \\
		Prompting & 19.36 & 6.88 & 3.59 & 3.67 & 16.09 & \textbf{8.93} & \textbf{10.81} &  \textbf{7.22} & \textbf{33.84} & 39.65 \\
		Reranking & 18.95 & 6.54 & 3.54 & 3.58 & 16.03 & 8.72 & 10.67 &  6.60 & 30.09 & 34.24\\
		\hline
		CE (ours) & \textbf{19.80} & \textbf{7.22} & \textbf{3.61} & \textbf{3.69} & \textbf{16.34} & 8.50 & 10.48 &  6.41 & 27.56 & \textbf{28.48} \\
		\Xhline{1pt}
	\end{tabular}
	\caption{Generation quality evaluation results on English dataset.}
	\label{table3}
\end{table*}

\begin{table}[t]
	\small
	\centering
	\renewcommand\arraystretch{1.2}
    \begin{tabular}{L{1.8cm}|C{0.7cm}|C{0.6cm}|C{0.6cm}|C{0.6cm}|C{0.6cm}}
	\Xhline{1pt}
	Metrics (\%) & K-C & ACC & P & R & F1 \\
	\Xhline{1pt}
	\multicolumn{6}{l}{\textbf{Training Transformer from scratch}} \\
	\hline
	K2S & 96.93 & 95.68 & 89.03 & 83.27 & 84.93 \\
	\hline
	Ctrl-decoding & 85.56 & 99.02 & 97.84 & 83.51 & 89.19 \\
	Prompting & 96.85 & 98.91 & 97.35 & 90.86 & 93.46 \\
	Reranking & 97.33 & 96.80 & 91.81 & 87.97 & 88.98 \\
	\hline
	CE (ours) & \textbf{98.00} & \textbf{99.10} & \textbf{98.09} & \textbf{92.84} & \textbf{94.96} \\
	\Xhline{1pt}
	\multicolumn{6}{l}{\textbf{Fine-tuning BART}} \\
	\hline
	K2S & 97.51 & 95.26 & 88.79 & 84.63 & 85.58 \\
	\hline
	Ctrl-decoding & 89.73 & \textbf{99.34} & \textbf{98.57} & 84.19 & 90.33 \\
	Prompting & 96.57 & 97.79 & 95.77 & 90.17 & 92.25 \\
	Reranking & 98.52 & 96.10 & 92.36 & 88.96 & 91.87 \\
	\hline
	CE (ours) & \textbf{98.68} & 99.13 & 98.54 & \textbf{93.72} & \textbf{95.77} \\
	\Xhline{1pt}
	\end{tabular}
	\caption{Satisfaction of controlling evaluation results on English dataset.}
	\label{table4}
\end{table}

\subsection{Evaluation Metrics}
% After building the benchmarking datasets, we need to introduce the evaluation metrics to assess the performance of the model. Lexical complexity controlled sentence generation is a CTG task. Therefore, it is concerned about not only the quality of the generated text but also the satisfaction with the controlled elements. We evaluate the performance from general text generation metrics and specific metrics for controlling the keywords and the complexity of words.

\subsubsection{Generated Quality} To evaluate the quality of generated text, we employ some automatic evaluate metrics in three aspects. 1) N-gram Similarity with References: we use \textbf{
BLEU} \cite{papineni2002bleu}, \textbf{METEOR} \cite{lavie2007meteor}, and \textbf{NIST} \cite{doddington2002automatic} evaluate the difference between generated texts and reference texts, which are commonly utilized in machine translation and text generation. 2) Diversity: We use 2-gram and 4-gram of \textbf{Entropy} \cite{zhang2018generating} and 1-gram and 2-gram of \textbf{Distinct} \cite{li2015diversity} to evaluate lexical diversity. 3) Fluency: Following \citet{zhang2020pointer, mcmc-xlnet}, to assess the fluency of generated sentences, we report the perplexity (\textbf{PPL}) over the test set using the pre-trained GPT-2 \cite{radford2019language} large model.
% \end{itemize}

\subsubsection{Satisfaction of Controlling} The control elements of lexical complexity controlled sentence generation have introduced in the Section \ref{define}. Our metrics are corresponding to the three constraints.
\begin{itemize}
\item \textbf{Keywords Constraint}. For this aspect, we introduce Keywords Constraint (\textbf{K-C}) satisfaction metric on word-level, which is computed using the percentage of the keywords contained in the generated sentences. The formular describe is as below:
\begin{equation}
K-C=\frac{1}{N}\sum\nolimits_{i=1}^{N}{\mathrm{count}_i^{C_1}}\big/m_i
\end{equation}
% \begin{equation}
% K-C=\frac{1}{N}\sum\nolimits_{i=1}^{N}{K_i^a}\big/{K_i}
% \end{equation}
where $N$ is the total number of samples in the test dataset, $\mathrm{count}_i^{C_1}$ is the number of keywords included in the generated sentence of the $i$-th sample, which satisfy the constraint of $C_1$, and $m_i$ is the number of the keywords of the input on the $i$-th sample. 
\item \textbf{Word Complexity Constraint}. The purpose of this metric is to calculate the Accuracy (\textbf{ACC}) of the words that meet the lexical complexity levels requirement in the generated sentence. As shown in the following formula:
\begin{equation}
ACC=\frac{1}{N}\sum\nolimits_{i=1}^{N}{\mathrm{count}_i^{C_2}}\big/{t_i}
\end{equation}
where $\mathrm{count}_i^{C_2}$ is the number of the words that satisfy the constraint $C_2$ of the $i$-th sample, and $t_i$ is the length of the generated sentence of the $i$-th sample. 
\item \textbf{Complexity Levels Constraint}. We propose three metrics to evaluate the satisfaction of the species of the required complexity levels. It is unreasonable that the ACC is still 100\% if given two complexity levels but the words of generated sentence only covers one of the levels. Thus we design the metrics of Precision (\textbf{P}), Recall (\textbf{R}), and \textbf{F1} to calcuate the saticfaction of complexity level constraint. The formular discribes are as follows:
% $$P=\frac{1}{N}\sum_{i=1}^{N}\frac{L_i^c}{L_i^g}$$
% $$R=\frac{1}{N}\sum_{i=1}^{N}\frac{L_i^c}{L_i^r}$$
% $$F1=\frac{1}{N}\sum_{i=1}^{N}\frac{2\frac{L_i^c}{L_i^g}\frac{L_i^c}{L_i^r}}{\frac{L_i^c}{L_i^g}+\frac{L_i^c}{L_i^r}}$$
\begin{equation}
P=\frac{1}{N}\sum\nolimits_{i=1}^{N}{\mathrm{count}_i^{C_3}}\big/{g_i}
\end{equation}

\begin{equation}
R=\frac{1}{N}\sum\nolimits_{i=1}^{N}{\mathrm{count}_i^{C_3}}\big/{n_i}
\end{equation}

% \begin{equation}
% \label{formular6}
% F1=\frac{1}{N}\sum\nolimits_{i=1}^{N}{2P_iR_i}\big/(P_i+R_i)
% \end{equation}
% where
% \begin{equation}
% \label{formular7}
% \begin{aligned}
% P_i&={\mathrm{count}_i^{C_3}}\big/{g_i} \\
% R_i&={\mathrm{count}_i^{C_3}}\big/{n_i}
% \end{aligned}
% \end{equation}
% then, substituting formula \ref{formular6} into \ref{formular7}, we can get:
\begin{equation}
\begin{aligned}
F1&=\frac{2}{N}\sum\nolimits_{i=1}^{N}{\mathrm{count}_i^{C_3}}\big/({n_i}+{g_i})
\end{aligned}
\end{equation}
where $\mathrm{count}_i^{C_3}$ is the number of the complexity levels satisfy the constraint $C_3$ of the $i$-th sample, $n_i$ is the number of the complexity levels given in the source  of the $i$-th sample, and $g_i$ is the number of the complexity levels of the generated sentence of the $i$-th sample.
\end{itemize}

\begin{table*}[thb!]
	\small
	\centering
	\renewcommand\arraystretch{1.2}
	\begin{tabular}{L{2cm}|R{0.7cm}R{0.7cm}|R{0.7cm}R{0.7cm}|C{1.6cm}|R{0.7cm}R{0.7cm}|R{0.7cm}R{0.7cm}|R{1cm}}
		\Xhline{1pt}
		\multirow{2.5}{*}{Metrics} & \multicolumn{2}{c|}{\ \ \ \ BLEU(\%)} & \multicolumn{2}{c|}{\ \ \ \ NIST(\%)} & \multirow{2.5}{*}{METEOR(\%)} & \multicolumn{2}{c|}{\ \ \ Entropy(\%)} & \multicolumn{2}{c|}{\ \ \ Distinct(\%)}  &  \multirow{2.5}{*}{PPL}  \\[3pt]
		& B-2 & B-4  & N-2 & N-4  &  & E-2 & E-4  & D-1 & D-2 &    \\[3pt]
		\Xhline{1pt}
		 \multicolumn{11}{l}{\textbf{Training Transformer from scratch}} \\
% 		\Xhline{1pt}
        \hline
        K2S & 13.92 & 4.17 & 2.73 & 2.76 & 15.00 & 8.83 & 10.20 &  8.60 & 37.70 & 48.32 \\
		\hline
		Ctrl-decoding & 12.84 & 3.57 & 2.48 & 2.50 & 13.70 & 8.70 & 10.30 &  6.08 & 34.90 & 224.59 \\
		Prompting & 13.90 & 3.81 & 2.70 & 2.73 & 14.35 & 8.53 & 10.05 &  7.47 & 33.35 & 45.61 \\
		Reranking & 15.46 & 5.37 &	\textbf{2.98} & \textbf{3.02} & 15.34 & 8.84 & 10.15 & 9.13 & 37.88 & 38.56 \\
% 		\Xhline{1pt}
        \hline
		CE (ours) & \textbf{15.69} & \textbf{6.27} & 2.91 & 2.94 & \textbf{16.04} & \textbf{9.28} & \textbf{10.58} & \textbf{10.68} & \textbf{47.71} & \textbf{34.53}  \\
		\Xhline{1pt}
        \multicolumn{11}{l}{\textbf{Fine-tuning BART}} \\
        \hline
		K2S & 14.97 & 4.39 & 3.08 & 3.10 & 16.56 & 8.60 & 10.06 & 9.91 & 37.13 & \textbf{21.76} \\
		\hline
		Ctrl-decoding & 12.54 & 3.71 & 2.38 & 2.55 & 14.04 & 8.73 & 10.25 & 9.96 & 37.85 & 129.86 \\
		Prompting & 16.81 & 5.47 & 3.15 & 3.17 & 16.24 & 8.69 & 10.13 & 10.04 & 38.33 & 31.75 \\
		Reranking & 16.53 & 6.42 & \textbf{3.29} & \textbf{3.36} & 16.61 & 8.81 & 10.08 & 10.15 & 38.96 & 53.47 \\
		\hline
		CE (ours) & \textbf{17.07} & \textbf{6.46} & 3.18 & 3.26 & \textbf{16.73} & \textbf{9.34} & \textbf{10.27} & \textbf{10.55} & \textbf{48.76} & 26.52 \\
		\Xhline{1pt}
	\end{tabular}
	\caption{Generation quality evaluation results on Chinese dataset.}
	\label{table5}
\end{table*}

\begin{table}[thb!]
	\small
	\centering
	\renewcommand\arraystretch{1.2}
    \begin{tabular}{L{1.8cm}|C{0.7cm}|C{0.6cm}|C{0.6cm}|C{0.6cm}|C{0.6cm}}
	\Xhline{1pt}
	Metrics (\%) & K-C & ACC & P & R & F1 \\
	\Xhline{1pt}
	\multicolumn{6}{l}{\textbf{Training Transformer from scratch}} \\
	\hline
	K2S & 87.36	& 92.74 & 85.40 & 68.40 & 73.75 \\
	\hline
	Ctrl-decoding & 71.83 & \textbf{99.96} & \textbf{99.96} & 61.79 & 74.73 \\
	Prompting & 85.54 & 98.88 & 97.79 & 80.23 & 86.88\\
	Reranking & 88.22 & 96.70 & 93.05 & 75.74 & 81.59 \\
	\hline
	CE (ours) & \textbf{89.61} & 98.87 & 97.49 & \textbf{88.80} & \textbf{92.17} \\
	\Xhline{1pt}
	\multicolumn{6}{l}{\textbf{Fine-tuning BART}} \\
	\hline
	K2S & 92.12 & 93.73 & 86.88 & 68.87 & 74.37 \\
	\hline
	Ctrl-decoding & 82.52 & \textbf{99.18} & \textbf{98.65} & 65.26 & 76.41 \\
	Prompting & 86.94 & 98.73 & 97.98 & 81.78 & 88.02 \\
	Reranking & 90.14 & 97.21 & 95.44 & 76.78 & 83.95 \\
	\hline
	CE (ours) & \textbf{92.58} & 99.07 & 97.91 & \textbf{89.34} & \textbf{92.85} \\
	\Xhline{1pt}
	\end{tabular}
	\caption{Satisfaction of controlling evaluation results on Chinese dataset.}
	\label{table6}
\end{table}

\section{Experiments}
Our experiments are based on the two datasets introduced in Section \ref{data-metric}. Besides the strong baselines of controlled decoding, prompting and reranking mentioned in Section \ref{baselines}, we generate the sentence by setting the keys as the input directly as the basic baseline (K2S). This baseline does not require complexity levels, which are just learnt from the data. Our evaluations include automatic evaluation and human evaluation. The automatic metrics have been introduced in the Section \ref{data-metric}. 
%We have introduced the datasets, main baselines and automatic evaluation metrics in previous sections. 

\subsection{Experimental Setup}
Our experimental setup contains two aspects:training from scratch and fine-tuning. From scratch training experiments are on the Transformer model \cite{Transformer}, which is the most widely used model in text generation. The fine-tuning experiments are on the pre-trained model of BART \cite{lewis2019bart}, which has superior generation ability. During inference, we run greedy decoding on all models for a fair comparison. We implement all models with the Fairseq library \footnote{https://github.com/pytorch/fairseq} and the BART pre-trained model is from HuggingFace Transformers library \footnote{https://github.com/huggingface/transformers} \cite{wolf2019huggingface}. All models are trained and tested on NVIDIA TITAN Xp GPU.

\subsubsection{From Scratch Training Setup}
We adopt the typical Transformer \cite{Transformer} as the model trained from scratch. We utilize a learning rate of 3e-4 and set the warming-up schedule with 4000 steps for training. We train our model for around 100 epochs. The optimization algorithm is Adam \cite{kingma2014adam}. We set the maximum number of input tokens as 8192, which is the same as transformer-based baselines. 
\subsubsection{Fine-tuning Setup}
We initialize our model with BART-base \cite{lewis2019bart}, which has comparable parameters to generation baselines. For generation baselines and our models, we use Adam \cite{kingma2014adam} with an initial learning rate of 1e-5 to update parameters for four epochs and choose the checkpoints with the lowest validation loss. We train our model for around 30 epochs. We set the maximum number of input tokens as 2048.

% \subsubsection{Baseline Setup}

\subsection{Baseline} \label{baselines}
% For the task of lexical complexity controlled example generation, we propose several strong baseline methods to compare. In previous works, the method of controlled decoding is extensively used in neural machine translation and text generation for lexical constraint. Prompting technique is also an arisen approach for controllable text generation. And re-ranking method is an traditional and effective method for different kinds of control in text generation and machine translation. We will introduce our baseline methods from these three aspects above.
\subsubsection{Controlled decoding}
We consider a strategy of controlled decoding \cite{dathathri2019plug} to realize the generated sentence consists of the words belonging to the given complexity levels. Since we know the words of the complexity level to be used in the sentence, we can restrict the words of the subset of the vocabulary to only be used in the decoding stage. The specific method is to set the probability of words outside the subset to zero so that they can meet the requirements of the word complexity level.
\subsubsection{Prompting}
Prompting is another feasible method for controlled text generation \cite{zou2021controllable}. Inspired by the prefix-tuning \cite{liang2021prefix}, which uses continuous vectors as prompts, we add the required complexity levels as the prefix for controlling in the input of the generation model. 
\subsubsection{Reranking}
Inspired by previous works \cite{ravaut2022summareranker, pandramish2020checkpoint}, we select the sentence that best meets the lexical complexity requirements from the N-best candidates. We take the score that is the sum of $ACC$ score and $F1$ score on the test reference hypothesis from this N-best list and choose the candidate that has the largest score. The detail of the re-ranking method is shown as the Algorithm \ref{alg:algorithm} in Appendix \ref{algorithm}.
% the iteration outputs are selected as the N-best list. It implies that for the last iteration, we have the corresponding N-best list for a input sequence.

\begin{table}[t]
	\small
	\centering
	\renewcommand\arraystretch{1.2}
    \begin{tabular}{L{1.8cm}|C{1.3cm}|C{1.3cm}|C{1.3cm}}
	\Xhline{1pt}
	Metrics (\%) & Semantics & Fluency & Diversity \\
	\Xhline{1pt}
	\multicolumn{4}{l}{\textbf{English dataset}} \\
	\hline
	Ctrl-decoding & 2.68 & 2.40  &  2.92  \\
	Prompting & \textbf{4.63} & 3.25 & 3.45 \\
	Reranking & 4.60 & 3.39 & 3.40  \\
	\hline
	CE (ours) & 4.62 & \textbf{3.82} & \textbf{3.54}  \\
	\Xhline{1pt}
	\multicolumn{4}{l}{\textbf{Chinese dataset}} \\
	\hline
	Ctrl-decoding & 3.89 & 2.82 &3.27 \\
	Prompting & 4.23 & 3.08 & 3.02 \\
	Reranking & 4.37 & 3.29 & 3.16 \\
	\hline
	CE (ours) & \textbf{4.57} & \textbf{3.80} & \textbf{3.71}  \\
	\Xhline{1pt}
	\end{tabular}
	\caption{Human evaluations for fine-tuning BART model on two datasets.}
	\label{table7}
\end{table}

\subsection{Experimental Results}
The experimental results on English dataset are shown in Table \ref{table3} and Table \ref{table4}. From the evaluation of generation quality in Table \ref{table3}, it can be seen that the method of complexity embedding has competitive results in different aspects, especially on fluency. In general, the CE method has better performance in the control of lexical complexity, especially on the metrics of R and F1. The method of controlled decoding has poor performance on PPL because it forces the distribution of the logits to concentrate on the words of given complexity levels in the decoding stage. This hard constraint pattern will impact the fluency of the generated sentences. But its performances on the metrics of ACC and P are better than other methods from Table \ref{table4}. The methods of prompting and reranking are two competitive baselines. The prompting method has better performance in the control of the word complexity because it has considered the word complexity levels in training. But the reranking method has better generation quality on the whole metrics of Table \ref{table3}. 

The experimental results on Chinese dataset are shown in Table \ref{table5} and Table \ref{table6}. We can draw similar conclusions from these two tables. Our approach performs well in terms of both text generation quality and lexical complexity control. The rerank approach outperforms prompt in all aspects of generation quality, both in terms of similarity to ground truth and in diversity and fluency, and even achieves the best NIST metrics for the Chinese dataset.

\subsection{More Analyses and Discussion}
The CE method we proposed has an excellent performance in controlling lexical complexity. The reason is that the CE method not only keeps the consistency of training and prediction but also considers the information of the complexity at the token level. Thus, it has more precise control of lexical complexity. And it also has competitive generation quality in the aspect of fluency and similarity with the reference. From the metrics of Entropy and Distinct, its diversity has a little poor performance in terms of the fine-tuning pattern on the English dataset.  We think the main reason is that the vocabulary of the English word complexity levels is less than which of the Chinese, so the token level restrictions of complexity embedding will impact the diversity of the sentences. The Chinese dataset, on the other hand, has a much larger coverage of voabulary with complexity and the dataset comes from the field of second language teaching, so the diversity of our model is better. It is worth noting that our CE method performs best in terms of lexical complexity control, especially the metrics of K-C, R, and F1, compared to the baseline model. This indicates that the CE method has higher coverage on complexity levels due to it takes into account the complexity of each word.

\subsection{Human Evaluation}
We conduct a human evaluation to further compare our model with the three baselines with fine-tuning the BART model on two datasets. For each model, we randomly select 200 generated sentences from the test set for each dataset and invite three annotators to label the sentences, who are postgraduates of the major in linguistics. To evaluate the quality of the sentences, annotators rate the sentences on three dimensions:  semantic consistency between the keywords and sentence; the fluency of the sentence; the diversity of the sentence \cite{zhang2020pointer}.  The score is range from 0 to 5. As shown in Table \ref{table7}, our method has better performance at the three aspects of human evaluation, especially the fluency and diversity. We give some real cases of two datasets in the Appendix \ref{case}.

\section{Related Work}
\subsection{Related Tasks}
\subsubsection{Lexical Constraint Text Generation}
Lexical constraint text generation is to generate a complete text sequence, given a set of keywords as constraints \cite{zhang2020pointer}. Previous works involve enhanced beam search \cite{DBA, hu2019improved} and the stochastic search methods \cite{zhang2020pointer, sha2020gradient}. Currently, Seq2Seq-based models such as Transformer and pre-trained models have been increased in generation with lexical constraint \cite{wang2021mention,liu2020kg, wang2021retrieval, fan2020enhanced, liu2021constrained}. But lexically constrained text generation is not able to control the complexity of words used in the generation, which is different from our work.
% Example generation is similar to lexically constrained text generation, which requires the given keywords are incorporated into the generated sentence. But lexically constrained text generation is not able to control the complexity of words used in the generation, which makes it hard to be used in real applications, such as children reading generation and language teaching.
\subsubsection{Text Readability Assess}
Research has shown that lexical complexity is also a crucial aspect of evaluating the complexity of a text for text readability assess task \cite{chakraborty2021simple}. In the relevant study of sentence-level readability, it is generally accepted that apart from sentence length, the most predictive indicator is the number of difficult words in the sentence \cite{weiss2022assessing}. In our work, we follow the definition and vocabulary of lexical complexity of text readability assess.
% The concept of readability at the sentence level can be related to the selection of appropriate vocabulary and sentences. We adopted the vocabulary scale from the Common European Framework of Reference for Languages (CEFR)  \footnote{https://www.englishprofile.org/wordlists/evp} which contains guidelines for the creation of teaching material, and the assessment of L2 proficiency as the standard for grading the complexity of vocabulary \cite{soproni2020common, chen2018automatic}. Because of the particularity of Chinese, we adopted the graded lexicon syllabus from Chinese Proficiency Grading Standards for International Chinese Language Education which has a complete rating system with six levels from low to high. 
\subsubsection{Text Simplification}
In text simplification field, lexical substitution, the replacement of complex words with simpler alternatives, is an integral part of sentence simplification and has been the subject of previous work \cite{alonzo2020automatic, nishihara2019controllable}. Differently, our work can strictly control the lexical complexity levels of the generated sentence, not only simplify the lexical complexity.
%Some methods in the text simplification task pay attention to word complexity which is an important factor for a sentence or document simplification. In these methods

\subsection{Related Methods}
\subsubsection{Controlled Decoding}
\citet{dathathri2019plug} use the gradients of an external discriminator to direct the generation of a pre-trained language model toward the target topic.  \citet{yang2021fudge} directly modifies the output probabilities of a language model using the output of a discriminator that determines whether the future text will contain the desired attribute. Different from the controlled decoding methods, our method considers the constraint of lexical complexity during both training and prediction.

%Similarly, \citet{krause2020gedi} use a contrastive strategy to soft-control language generation.
% \citet{pascual2021plug} proposed a plug-and-play decoding method for controlled language generation that is so simple and intuitive. It can be described in a single sentence: given a topic or keyword, we add a shift to the probability distribution over our vocabulary towards semantically similar words.

\subsubsection{Prompting}
The prompting method has emerged as a new way to perform natural language processing by conditioning on extra information.
\citet{brown2020language} propose to use a task description and a few examples to adapt the GPT-3 model to downstream tasks, which is referred to as in-context learning. Their prompts are manually designed. \citet{gao2020making} present LM-BFF for automatic prompts generation. \citet{liang2021prefix} propose prefix-tuning, which uses continuous vectors as prompts. Compared to the prompting method, our method fuses more fine-grained information on lexical complexity in model training.

\subsubsection{Reranking}
The reranking approach has been proved to have excellent performance in machine translation \cite{pandramish2020checkpoint, wang2007reranking} and text generation \cite{ravaut2022summareranker}. The reranking method rescores the n-best candidates through a model or a function and selects the highest scoring candidate as the final prediction \cite{imamura2017ensemble}. Unlike the reranking method, our method do not need to process the outputs after decoding.  

\section{Conclusions}
To summarize, we introduce a new task of lexical complexity controlled sentence generation, where word complexity must be strictly controlled in generating. To promote the development of this task, we develop two datasets and four metrics for the controlled element. In this paper, we also develop a series of alternate solutions for this task and propose a novel method based on complexity embedding to obtain better control of lexical complexity in a generation. Our results indicate that the complexity embedding method has better performance in controlling the lexical complexity and competitive generation quality.

% \section*{Acknowledgements}
% This document has been adapted by Jordan Boyd-Graber, Naoaki Okazaki, Anna Rogers from the style files used for earlier ACL, EMNLP and NAACL proceedings, including those for
% EACL 2023 by Isabelle Augenstein and Andreas Vlachos,
% EMNLP 2022 by Yue Zhang, Ryan Cotterell and Lea Frermann,
% ACL 2020 by Steven Bethard, Ryan Cotterell and Rui Yan,
% ACL 2019 by Douwe Kiela and Ivan Vuli\'{c},
% NAACL 2019 by Stephanie Lukin and Alla Roskovskaya, 
% ACL 2018 by Shay Cohen, Kevin Gimpel, and Wei Lu, 
% NAACL 2018 by Margaret Mitchell and Stephanie Lukin,
% Bib\TeX{} suggestions for (NA)ACL 2017/2018 from Jason Eisner,
% ACL 2017 by Dan Gildea and Min-Yen Kan, NAACL 2017 by Margaret Mitchell, 
% ACL 2012 by Maggie Li and Michael White, 
% ACL 2010 by Jing-Shin Chang and Philipp Koehn, 
% ACL 2008 by Johanna D. Moore, Simone Teufel, James Allan, and Sadaoki Furui, 
% ACL 2005 by Hwee Tou Ng and Kemal Oflazer, 
% ACL 2002 by Eugene Charniak and Dekang Lin, 
% and earlier ACL and EACL formats written by several people, including
% John Chen, Henry S. Thompson and Donald Walker.
% Additional elements were taken from the formatting instructions of the \emph{International Joint Conference on Artificial Intelligence} and the \emph{Conference on Computer Vision and Pattern Recognition}.

\bibliography{anthology,custom}
\bibliographystyle{acl_natbib}

\appendix

\section{Complexity Embedding Id} \label{EmbID}

The English words have six levels. And the Chinese words have seven levels (Diff 1-7). We give the design of the complexity embedding id for this two language in the table \ref{table2}. Note that, if a word is out of the complexity level vocabulary, its complexity is ``$\langle out \rangle$'' which is mapping into id 7 in English corpus and 8 in Chinese corpus. In addition, the special tokens such as ``$\langle s \rangle$'' ``$\langle pad \rangle$" "$\langle \verb|\|s \rangle$'' ``$\langle unk \rangle$'' are the common meaning in data preprocessing for model training.

\begin{table}[h]
\small
\centering
\renewcommand\arraystretch{1.2}
\begin{tabular}{C{1.5cm}C{1.5cm}|C{1.5cm}C{1.5cm}}
	\Xhline{1pt}
	\multicolumn{2}{c|}{English} & \multicolumn{2}{c}{Chinese} \\
	\Xhline{0.5pt}
	Token & Id & Token & Id \\
	\Xhline{0.5pt}
    Punctuation & 0 & Punctuation & 0 \\
    A1-C2 & 1-6 & Diff 1-7 & 1-7 \\
    $\langle out \rangle$ & 7 & $\langle out \rangle$ & 8 \\
    $\langle sep \rangle$ & 8 & $\langle sep \rangle$ & 9 \\
    $\langle s \rangle$ & 8 & $\langle s \rangle$ & 9 \\
    $\langle pad \rangle$ & 8 & $\langle pad \rangle$ & 9 \\
   $ \langle \verb|\|s \rangle$ & 8 & $ \langle \verb|\|s \rangle$ & 9 \\
   $ \langle unk \rangle$ & 8 &  $ \langle unk \rangle$ & 9 \\
	\Xhline{1pt}
\end{tabular}
\caption{Complexity Embedding Id.}
\label{table2}
\end{table}

\section{Details of Datasets Construction} \label{dataDetails}
\subsection{English Dataset}
We adopt the English word complexity levels in the Common European Framework of Reference for Languages (CEFR) \footnote{https://www.englishprofile.org/wordlists/evp} which is divided into six complexity levels (A1, A2, B1, B2, C1, and C2). First, we need to restrict the words in the corpus to ensure most of the words are in the complexity level vocabulary. Then, we need to extract keywords from the sentences. In this process, we command the number of keywords is related to the length of the sentence, and the number of keywords is between 1 to 5. Finally, we obtain the complexity information of each sentence through the complexity level vocabulary. The English raw corpus is collected from the monolingual English News dataset in ACL2019 WMT. We select those sentences which have 90\% words in the complexity level vocabulary of CEFR. After the processes mentioned above, we get 199k samples in the English corpus, and we split the train, validation and test dataset as shown in the Table \ref{tabel1}.
\subsection{Chinese Dataset}
The word complexity levels in Chinese Proficiency Grading Standards for International Chinese Language Education (CPGS) \footnote{http://www.chinesetest.cn} is divided into six complexity levels (1 to 7). The Chinese raw corpus is collected from 500 textbooks for Chinese learners. These textbooks contain two types of text: essay and dialogue. We split these texts into sentences and throw away those short sentences. If the raw text is a dialogue, after splitting, we need to remove the speaker's name to guarantee it is a proper sentence. Then, we command the number of keywords is related to the length of the sentence, and the number of keywords is between 1 to 5.  After the processes mentioned above, we get 156k samples in the Chinese corpus, as shown in the Table \ref{tabel1}.

\section{Algorithm of Reranking} \label{algorithm}

The algorithm is the detail of reranking method. We select the sentence that best meets the lexical complexity requirements from the N-best candidates, and $N=10$. On the test set, We take the sum of $ACC$ score and $F1$ score. The, we choose the candidate that has the largest score.

\renewcommand{\algorithmicrequire}{\textbf{Input:}}
\renewcommand{\algorithmicensure}{\textbf{Output:}}

% \begin{algorithm}[h]
% \caption{Reranking Method}
% \label{alg:algorithm}
% \textbf{Input}: Generated $n$ best candidate sentences $H=(h_0, h_1, h_2,..., h_{n-1})$ for given keywords and $n=10$; \textbf{Output}: Sentence having highest score 
% \begin{algorithmic}[1] %[1] enables line numbers
% \STATE Let $score=0$ \\
% \FOR{ each sentence $h_j$ in $H$}
% \STATE $ACC = F_{acc}(h_j)$
% \STATE $F1 = F_{f1}(h_j)$
% %\IF {conditional}
% \STATE $score_j=ACC+F1$
% %\ELSE
% \IF {$score_j>score$}
% \STATE $score=score_j$
% \STATE $ret=h_j$
% \ENDIF
% \ENDFOR
% \STATE \textbf{return} $ret$
% \end{algorithmic}
% \end{algorithm}
\begin{algorithm}[!h]
 \begin{algorithmic}[1]
 \caption{Reranking Method}
 \label{alg:algorithm}
 \Require Generated $n$ best candidate sentences $H=(h_0, h_1, h_2,..., h_{n-1})$ for given keywords and $n=10$
 \Ensure Sentence having highest score
 
 \State Let $score=0$ 
  \For{ each sentence $h_j$ in $H$}
  \State $ACC = F_{acc}(h_j)$
  \State $F1 = F_{f1}(h_j)$
  %\IF {conditional}
  \State $score_j=ACC+F1$
  %\ELSE
  \If {$score_j>score$}
   \State $score=score_j$
   \State $ret=h_j$
  \EndIf
  \EndFor
 \State \textbf{return} $ret$
 \end{algorithmic}
\end{algorithm}
\section{Case Study} \label{case}
We choose some cases of the fine-tuning pattern from two datasets. The English cases are in the Table \ref{tabel8}, and the Chinese cases are in the Table \ref{tabel9}. In both tables, the required keywords as well as appearing in the sentences are shown in blue font, and certain given grades as well as words actually appearing in the sentences for the corresponding grade are shown in red font.

\begin{table*}[htb!]
	\small
	\centering
	\renewcommand\arraystretch{1.3}
	\begin{tabular}{L{1.8cm}|L{13cm}}
		\Xhline{1pt}
		\multicolumn{2}{l}{\textbf{Keywords:}  \textcolor{blue}{refuse}, \textcolor{blue}{accept}, \textcolor{blue}{country} ; \ \ \ \textbf{Complexity Levels:}  A1, A2, B1, \textcolor{red}{C1}}
		 \\
		%\Xhline{1pt}
		\Xhline{1pt}
		Ctrl-decoding & I \textcolor{blue}{refuse} to \textcolor{blue}{accept} that this is a matter of time for the \textcolor{blue}{country}, but I do not accept it. \\
		\hline
		Prompting  & I \textcolor{blue}{refuse} to \textcolor{blue}{accept} that I am the only one who has been elected to the \textcolor{blue}{country}, but I am not. \\
		%\Xhline{1pt}
		\hline
		Reranking & I \textcolor{blue}{refuse} to \textcolor{blue}{accept} that this is the best thing I can do for the \textcolor{blue}{country}, " he said. \\
		\Xhline{1pt}
		CE (ours) & I \textcolor{blue}{refuse} to {\textcolor{blue}{accept}} that this is a {\textcolor{blue}{country}} that is going to be a place where people are going to be \textcolor{red}{unsafe}. \\
		%\Xhline{1pt}
		\hline
		
		\hline
		\multicolumn{2}{l}{\textbf{Keywords:}  \textcolor{blue}{public}, \textcolor{blue}{maintain}, \textcolor{blue}{lose} ; \ \ \ \textbf{Complexity Levels:}  A1, A2, B1, \textcolor{red}{B2}}
		 \\
		\Xhline{1pt}
		Ctrl-decoding & I do not think the \textcolor{blue}{public} will \textcolor{blue}{maintain} the power to \textcolor{blue}{lose}, but I do not think it will. \\
		\hline
		Prompting  & The \textcolor{blue}{public} will \textcolor{blue}{maintain} the public's ability to lose, and the public will not \textcolor{blue}{lose}, and they will not lose. \\
		%\Xhline{1pt}
		\hline
		Reranking & I don't want to be in \textcolor{blue}{public}, but I don't want to \textcolor{blue}{maintain} my weight and \textcolor{blue}{lose}. \\
		\Xhline{1pt}
		CE (ours) & The \textcolor{blue}{public} must \textcolor{blue}{maintain} their \textcolor{red}{faith} and not \textcolor{blue}{lose} , and we will continue to do everything we can to protect them. \\
		\hline
		
		\hline
		\multicolumn{2}{l}{\textbf{Keywords:}  \textcolor{blue}{football}, \textcolor{blue}{Leicester}, \textcolor{blue}{City}, \textcolor{blue}{magical} ; \ \ \ \textbf{Complexity Levels:}  A1, A2, B1, B2, \textcolor{red}{C2}}
		 \\
		\Xhline{1pt}
		Ctrl-decoding & I think \textcolor{blue}{football} is a great way to play for the game and to be able to play for the best of the game against the game against the game against the game and the way we play against the game against the game against the game against the game is not the same, but the way we are \textcolor{blue}{magical}. \\
		\hline
		Prompting  & I think \textcolor{blue}{football} is going to be the best of \textcolor{blue}{Leicester} \textcolor{blue}{City}, but I think it's a \textcolor{blue}{magical} moment for us to have to play with. \\
		%\Xhline{1pt}
		\hline
		Reranking & This is a \textcolor{blue}{football} game for \textcolor{blue}{Leicester} \textcolor{blue}{City}, and it's a \textcolor{blue}{magical} moment for us to be part of the game and be a part of it. \\
		\Xhline{1pt}
		CE (ours) & It's a \textcolor{blue}{football} club that has been around for a long time and to see \textcolor{blue}{Leicester} \textcolor{blue}{City} win the \textcolor{red}{title} is \textcolor{blue}{magical}. \\
		\Xhline{1pt}
	\end{tabular}
	\caption{ Generated examples from the English dataset.}
	\label{tabel8}
\end{table*}

\begin{table*}[htb!]
	\small
	\centering
	\renewcommand\arraystretch{1.3}
	\begin{tabular}{L{1.8cm}|L{13cm}}
		\Xhline{1pt}
		%\hline
		\multicolumn{2}{l}{\textbf{Keywords:}  \chinese{\textcolor{blue}{生活}} ; \ \ \ \textbf{Complexity Levels:}  1, 2, 3, \textcolor{red}{7}}
		 \\
		%\Xhline{1pt}
		\Xhline{1pt}
		Ctrl-decoding & \chinese{\textcolor{blue}{生活}中，我最喜欢的就是吃面。} \\
		\hline
		Prompting  & \chinese{\textcolor{blue}{生活}中的一切都是为了生活。} \\
		%\Xhline{1pt}
		\hline
		Reranking & \chinese{我们的\textcolor{blue}{生活}很美好，也很快乐。} \\
		\Xhline{1pt}
		CE (ours) & \chinese{没有工作压力的\textcolor{blue}{生活}多让人\textcolor{red}{羡慕}啊！} \\
		%\Xhline{1pt}
		\hline
		
		\hline
		%\hline
		\multicolumn{2}{l}{\textbf{Keywords:}  \chinese{\textcolor{blue}{春天}}, \chinese{\textcolor{blue}{姑娘}} ; \ \ \ \textbf{Complexity Levels:}  1, 2, 3}
		 \\
		%\Xhline{1pt}
		\Xhline{1pt}
		Ctrl-decoding & \chinese{\textcolor{blue}{春天}来了，\textcolor{blue}{姑娘}又来了，他还在那儿等着我呢。} \\
		\hline
		Prompting  & \chinese{\textcolor{blue}{春天}来了，\textcolor{blue}{姑娘}又来了，他的眼睛也变得很美丽。} \\
		%\Xhline{1pt}
		\hline
		Reranking & \chinese{\textcolor{blue}{春天}到了，一个年轻的\textcolor{blue}{姑娘}爱上了他。} \\
		\Xhline{1pt}
		CE (ours) & \chinese{\textcolor{blue}{春天}来了，我认识了一位美丽的\textcolor{blue}{姑娘}，他非常喜欢这个姑娘。} \\
		%\Xhline{1pt}
		\hline
		
		\hline
		%\hline
		\multicolumn{2}{l}{\textbf{Keywords:}  \chinese{\textcolor{blue}{喜欢}} ; \ \ \ \textbf{Complexity Levels:}  1, \textcolor{red}{7} }
		 \\
		%\Xhline{1pt}
		\Xhline{1pt}
		Ctrl-decoding & \chinese{我\textcolor{blue}{喜欢}吃中国菜，我\textcolor{blue}{喜欢}吃中国菜。} \\
		\hline
		Prompting  & \chinese{我\textcolor{blue}{喜欢}看中国电影。
}\\
		%\Xhline{1pt}
		\hline
		Reranking & \chinese{你\textcolor{blue}{喜欢}什么样的女孩子呢？}\\
		\Xhline{1pt}
		CE (ours) & \chinese{我最\textcolor{blue}{喜欢}\textcolor{red}{逗}它们玩儿。} \\
		%\Xhline{1pt}
		\hline
		
		\hline
		%\hline
		\multicolumn{2}{l}{\textbf{Keywords:}  \chinese{\textcolor{blue}{别人}}, \chinese{\textcolor{blue}{可能}}, \chinese{\textcolor{blue}{意思}} ; \ \ \ \textbf{Complexity Levels:}  1, 2}
		 \\
		%\Xhline{1pt}
		\Xhline{1pt}
		Ctrl-decoding & \chinese{\textcolor{blue}{别人}看我懂了，我觉得我有\textcolor{blue}{可能}去做一个人，\textcolor{blue}{意思}是要看他。} \\
		\hline
		Prompting  & \chinese{\textcolor{blue}{别人}\textcolor{blue}{可能}不会说，如果你觉得自己可能有可能，你可能会觉得自己是个很难的\textcolor{blue}{意思}。}\\
		%\Xhline{1pt}
		\hline
		Reranking & \chinese{如果\textcolor{blue}{别人}问你一个问题，你的\textcolor{blue}{意思}是什么？} \\
		\Xhline{1pt}
		CE (ours) & \chinese{\textcolor{blue}{别人}\textcolor{blue}{可能}不知道你的\textcolor{blue}{意思}，你要做我喜欢的，要我愿意跟别人说。} \\
		%\Xhline{1pt}
		\Xhline{1pt}
		
	\end{tabular}
	\caption{ Generated examples from the Chinese dataset.}
	\label{tabel9}
\end{table*}

\end{document}